\documentclass{article}
\usepackage[margin=1.1in]{geometry}
\usepackage[colorlinks=true,linkcolor=green,citecolor=blue]{hyperref}%
\usepackage{todonotes}
\usepackage[round]{natbib}
\usepackage{microtype}
\usepackage{graphicx}
\usepackage{grffile}
\usepackage{subfigure}
\usepackage{booktabs}
\usepackage{hyperref}
\usepackage{algorithmic,algorithm}
\usepackage{amsmath}
\usepackage{amsfonts}
\usepackage{amssymb}
\usepackage{lmodern} 
\usepackage{diagbox}
\usepackage{dsfont,bm,bbm}

\newcommand{\cR}{\mathcal{R}}

\newcommand{\cU}{\mathcal{U}}

\newcommand{\bE}{\mathbb{E}}
\newcommand{\bP}{\mathbb{P}}

\newcommand{\argmax}[1]{\underset{#1}{\mathrm{argmax }}}

\renewcommand{\epsilon}{\varepsilon} 
 
\newtheorem{proposition}{Proposition}
\newtheorem{lemma}{Lemma}

\newcommand{\indic}{\mathbbm{1}}

\title{A High Performance, Low Complexity Algorithm for Multi-Player Bandits Without Collision Sensing Information}
\author{Cindy Trinh \thanks{CentraleSupelec, L2S, France} \,and Richard Combes \thanks{CentraleSupelec, L2S, France}}
\begin{document}
\maketitle
\begin{abstract}
Motivated by applications in cognitive radio networks, we consider the decentralized multi-player multi-armed bandit problem, without collision nor sensing information. We propose Randomized Selfish KL-UCB, an algorithm with very low computational complexity, inspired by the Selfish KL-UCB algorithm, which has been abandoned as it provably performs sub-optimally in some cases. We subject Randomized Selfish KL-UCB to extensive numerical experiments showing that it far outperforms state-of-the-art algorithms in almost all environments, sometimes by several orders of magnitude, and without the additional knowledge required by state-of-the-art algorithms. We also emphasize the potential of this algorithm for the more realistic dynamic setting, and support our claims with further experiments. We believe that the low complexity and high performance of Randomized  Selfish KL-UCB makes it the most suitable for implementation in practical systems amongst known algorithms.
\end{abstract}

\section{Introduction}
\label{introduction}

The multi-armed bandit problem is an online decision-making problem introduced by \cite{thompson} nearly a century ago in the context of clinical trials. It has then been widely studied for its numerous applications, for example in recommendation systems, advertising or hyperparameter tuning.

We are in particular interested in the multi-player version of the multi-armed bandit problem where at each time step, multiple players choose among a common set of arms. If two players choose the same arm, they collide and both of them receive a null reward. If a player is the only one to choose an arm, they receive a reward sampled from a Bernoulli distribution. Let us formally state the model.

\subsection{Model and Assumptions}

\paragraph{Multi-Player Bandits} We consider an $M$-player $K$-armed bandit problem with $M \le K$, parameterized by $\bm\mu = (\mu_1, \dots, \mu_K)$. The ``true'' reward of arm $k=1,...,K$ at time $t=1,...,T$ is denoted by $X_k(t)$, and we assume that $(X_k(t))_{t \ge 1}$ is i.i.d. Bernoulli distributed with expected value $\bE[X_k(t)] = \mu_k$. At time $t=1,...,T$, player $m=1,...,M$ chooses an arm $\pi_m(t) \in \{1,...,K\}$ based on their past observations. If two players choose the same arm, we say that a collision occurs and both of them receive a reward of $0$. Formally, we define the collision indicator variable
$$
    \eta_{k}(t) = \mathbbm{1} \Bigg\{ \# \{m=1,...,M: \pi_m(t) = k\} \geq 2 \Bigg\} 
$$
so that $\eta_{k}(t) = 1$ if at least two distinct players have chosen arm $k$, leading to a collision, and $\eta_{k}(t) = 0$ otherwise.

\paragraph{Rewards and Information Structure}

The reward obtained by player $m=1,...,M$ can then be written as
$$
    r_m(t) = X_{\pi_m(t)}(t)\Big(1 - \eta_{\pi_m(t)}(t)\Big),
$$
so that if they collide with another player, they get a $0$ reward and otherwise they get the reward of the selected arm $\pi_{m}(t)$ which is $X_{\pi_m(t)}(t)$.

Informally, the multi-player multi-armed bandit problem is simply an extension of the classical single-player multi-armed bandit problem, where several users simultaneously explore the arms in a distributed manner, subject to collisions.

In this work we consider no collision nor sensing information, which is the setting in which one gets the minimal amount of information. Namely, player $m$ observes solely its reward $r_m(t)$. They do not observe the channel reward $X_{\pi_m(t)}(t)$, nor do they observe the collision indicator $\eta_{\pi_m(t)}(t)$ or the rewards obtained by other players. In short the problem is fully distributed. We also assume that initially players have no information whatsoever about the mean reward of arms $\bm\mu$, no information about the number of players $M$, and no information about their index $m$, so that all players must behave symmetrically.

\paragraph{Optimal Policy and Regret}

In order to maximize the total reward, players must find the $M$ best arms, and assign one distinct arm to each player in order to avoid collisions, which would yield a total expected reward of $\sum_{m=1}^{M} \mu_{(m)}$,
where $\mu_{(1)} \geq ... \geq \mu_{(K)}$ are the sorted expected rewards of each arm $\mu_1,...,\mu_{K}$. We define the regret
$$
 R_{\bm\mu}(T) = \sum_{t=1}^T \left(\sum_{m=1}^M \mu_{(m)}  - \sum_{m=1}^M \bE [r_{\pi_m(t)}(t)] \right) 
$$
which is the difference in terms of cumulative expected reward between an oracle that knows $\bm\mu$ and acts optimally versus that of the algorithm considered.

\subsection{Related Work}

\paragraph{Centralized setting}
When players' decisions are managed by a central controller, we say the system is centralized, and this amounts to the multiple-play bandits, where the controller chooses a set of $M$ arms at each time step. This model was introduced by \cite{anantharam} and further studied by \cite{komiyama15}. The high cost of a central controller in cognitive radio networks applications however motivated \cite{liuzhao} to introduce the more interesting but more difficult decentralized setting, where players can only observe their own actions and received rewards.

\paragraph{Decentralized setting, with collision/sensing information}
Since then, the decentralized setting assuming collision information (knowledge of the collision indicator $\eta_{\pi_m(t)}(t)$) and/or sensing information (knowledge of $X_{\pi_m(t)}$) has been well explored \citep{liuzhao, anandkumar, rosenski, besson_multiplayer_revisited, boursier_sicmmab}.  Recently, \cite{boursier_sicmmab} introduced SIC-MMAB, an algorithm which carefully leverages collisions between players in order to communicate. By doing so, they prove that its regret matches asymptotically the lower bound of the centralized problem, up to a universal constant.

\paragraph{Decentralized setting, without collision nor sensing information}
The decentralized setting without collision nor sensing information has been much less investigated. It was first introduced by \cite{bonnefoi}. They proposed the Selfish UCB algorithm and its application to IoT networks showed promising experimental results. Unfortunately, \cite{besson_multiplayer_revisited} then conjectured that with constant probability, it may incur linear regret, a negative result which was further confirmed by \cite{boursier_sicmmab}. Meanwhile, \cite{lugosi} provided the first two algorithms proven to achieve a logarithmic regret. \cite{boursier_sicmmab} followed with SIC-MMAB2, an adapted version of SIC-MMAB, which also achieves a logarithmic regret. Building upon their work, \cite{shi_ec_sic} proposed EC-SIC, an algorithm which improves the efficiency of the communication phase of SIC-MMAB2 by cleverly using an error correction code to transmit full statistics of the players. In Table~\ref{related_work} we report known  regret upper bounds for these algorithms.

In the decentralized setting without collision and sensing information, the question of whether an algorithm can reach similar performance as in the centralized setting remains unanswered. For the simplest case where $M=2$ and $K =3$, \cite{bubeck} proposed a collision avoiding algorithm which achieves a problem independent regret of $O(\sqrt{T\log T})$. Note however that their setting is slightly different as it is cooperative, meaning that the two players have assigned roles at the beginning of the game. Their lower bound on the full-information feedback model also suggests that the $\log T$ term is necessary for the bandit feedback model.

\begin{table*}[t] \label{related_work}
\caption{Existing algorithms for decentralized, no collision/sensing information, no cooperative algorithm }
\vskip 0.15in
\begin{small}
\begin{tabular}{lccr}
\toprule
Algorithm & Required Knowledge & Asymptotic Upper Bound   \\
\midrule
Algorithm 1  & $M$ & $O\left(\frac{MK\log T}{( \mu_{(M)} - \mu_{(M+1)} )^2}\right)$ \\
 \cite{lugosi}  & & \\
Algorithm 2  & $\mu_{(M)}$ & $O\left(\frac{K^2M \log^2 T }{\mu_{(M)}} + KM \min (\sqrt{T \log T}, \frac{\log T}{\mu_{(M)}- \mu_{(K)}}  \right)$ \\
 \cite{lugosi}  & & \\
SIC-MMAB2  & $\mu_{(K)}$ & $ 
 O\left(\sum_{k>M} \min \lbrace \frac{M\log T}{\mu_{(M)} - \mu_(k)}, \sqrt{MT \log T} \rbrace + \frac{M K^2}{\mu_{(K)}} \log T\right)$  \\  
  \cite{boursier_sicmmab}  & & \\
EC-SIC  & $\Delta$, $\mu_{(K)}$ & $O\left(\sum_{k>M}  \frac{\log T}{\mu_{(M)} - \mu_{(k)}} + \left(\frac{M^2 K}{E(\mu_{(K)})} \log \frac{1}{\Delta} + \frac{MK}{\mu_{(K)}} \right) \log T \right) $ \\ 
  \cite{shi_ec_sic}  & & \\
\bottomrule
\end{tabular}
\end{small}
\vskip -0.1in
\end{table*}

\paragraph{Dynamic setting} An even more interesting setting for real-world applications is the dynamic setting, where the number of players $M$ is no longer fixed: players can leave or enter the game. Under the collision information assumption, \cite{avner2014concurrent} propose the MEGA algorithm, and show that it is robust when a player leaves the game. Later on, in the setting where players can leave the game only after a specific time, \cite{rosenski} propose the Dynamic Musical Chairs algorithm which consists in resetting the Musical Chairs algorithm at a certain frequency. For the dynamical setting without collision nor sensing information, the literature is still scarce at the moment. \cite{boursier_sicmmab} proposed an algorithm with logarithmic regret, DYN-MMAB, under quasi-asynchronicity, that is, the hypothesis that the players can enter but cannot leave the game. When players are allowed to leave, they suggest to generalize their algorithm by resetting it.

\subsection{Contributions}

A drawback of current state-of-the-art algorithm in the decentralized setting without collision nor sensing information is that they all assume the unrealistic knowledge of certain parameters of the environment such as the number of players $M$ \citep{lugosi}, the mean reward of the $M$-th best channel $\mu_{(M)}$ \citep{lugosi}, the gap $\Delta = \mu_{(M)}-\mu_{(M+1)}$ and/or a lower bound $\mu_{\min}$ on $\mu_{(K)}$ \citep{boursier_sicmmab, shi_ec_sic}, which are usually unknown to the users in real-world applications.

We propose Randomized Selfish KL-UCB, an algorithm derived from Selfish KL-UCB, which does not rely on such unrealistic assumptions. This algorithm also does not suffer from the negative results of Selfish KL-UCB stressed by \cite{besson_multiplayer_revisited, boursier_sicmmab}, and we show through extensive experiments that it performs far better than state-of-the-art algorithms in almost all environments, except some edge cases (Section~\ref{comparison_sota}), where it still seems to incur a logarithmic regret. Moreover, our experiments reveal that, in some environments, the performance of Randomized Selfish KL-UCB is even better than state-of-the-art algorithms which assume collision or sensing information such as SIC-MMAB \citep{boursier_sicmmab}, and MCTopM \citep{besson_multiplayer_revisited} (Section~\ref{sec:comparison_wo_coll_info}).

For the dynamic setting, we carry out experiments which also emphasize the potential of Randomized Selfish KL-UCB. 
Under quasi-asynchronicity assumption, our experiments show that Randomized Selfish KL-UCB outperforms by far DYN-MMAB.

Furthermore, we propose a new, more realistic dynamic setting, where players can enter and leave at any moment. For this setting, since no algorithm exists to the best of our knowledge, we compare our algorithm to a simple Musical Chairs \cite{rosenski} for a baseline, and show that again, Randomized Selfish KL-UCB is very promising.

All code used for conducting experiments is publicly available at \url{https://github.com/ctrnh/multi_player_multi_armed_bandit_algorithms}.

\section{Proposed Algorithm}

We now highlight the proposed algorithm, the rationale behind its construction and list some of the shortcomings of the state-of-the-art algorithms.

\subsection{Computation of Statistics}
 For player $m=1,...,M$ and arm $k=1,...,K$ we define
$$
    N_{m,k}(t) = \sum_{s=1}^{t-1} \indic \{ \pi_m(s) = k \},
$$
the number of times player $m$ has selected arm $k$ between time step $1$ and $t-1$, as well as
$$
    \hat{\mu}_{m,k}(t) = {1  \over  \max( N_{m,k}(t),1)  } \sum_{s=1}^{t-1} r_{m}(s) \indic \{ \pi_m(s) = k \},
$$
the average empirical reward obtained by player $m$ from arm $k$ between time step $1$ and $t-1$. Note that for all $k=1,...,K$, both $N_{m,k}(t)$ and $\hat{\mu}_{m,k}(t)$ are available to player $m$ at time $t$ based on the model assumptions. We will use those two statistics in order to design our algorithms.

\subsection{The Selfish KL-UCB Algorithm}

An algorithm proposed by \cite{besson_multiplayer_revisited} called Selfish KL-UCB is that each player $m=1,...,M$ chooses the arm 
$$
    \pi_m(t) \in \arg\max_{k=1,...,K} \{ b_{m,k}(t) \}
$$
maximizing the KL-UCB index defined as
$$
    b_{m,k}(t) = \max \{q \in [0,1]: N_{m,k}(t) d(  \hat{\mu}_{m,k}(t) , q ) \le f(t) \}
$$
where $f(t) = \log t + c \log\log t$, with $c\ge 0$ (in practice, one usually simply sets $c=0$) and 
$$
d(\mu ,  \lambda) = \mu \log {\mu \over \lambda} + (1-\mu) \log {1-\mu \over 1 - \lambda}
$$ 
is the Kullback-Leibler divergence between Bernoulli distributions with means $\mu$ and $\lambda$. The pseudo code for Selfish KL-UCB is stated as Algorithm~\ref{alg:SelfishKLUCB}.

As its name indicates, Selfish KL-UCB is a straightforward extension to the multi-player setting of the KL-UCB algorithm \cite{klucb}, an optimistic algorithm which is provably asymptotically optimal in the single-player setting. It is called ``Selfish'' since each player acts as if other players did not exist and attempts to play optimally in the single player setting.

While Selfish KL-UCB is both conceptually simple and elegant, there exists cases in which its regret is linear, as proven by \cite{boursier_sicmmab} and stated in proposition~\ref{prop:selfishkl} \footnote{To be more accurate, the authors of  \cite{boursier_sicmmab} analyze Selfish UCB1 which is simply Selfish KL-UCB using the UCB1 index instead of the KL-UCB index. Experimentally, both Selfish UCB1 and Selfish KL-UCB exhibit the same problematic behaviour.}. Selfish KL-UCB has an all-or-nothing behaviour: on some sample paths it performs very well, and on some of them a subset of players simply collide without an end, causing linear regret.
\begin{proposition}\label{prop:selfishkl}
    There exists $K$, $M$ and $\bm\mu \in [0,1]^K$ such that under Selfish UCB1:
    $$
        \lim \inf_{T \to \infty} {R_{\bm\mu}(T) \over T} > 0 
    $$
    i.e. the regret grows linearly.
\end{proposition}

\subsection{Proposed Algorithm: Randomized Selfish KL-UCB}

The reason why Selfish KL-UCB performs poorly on some sample paths is due to its symmetry. Indeed, consider two players $m \ne m'$ who, at time $t \in [T]$, have the same observations, that is $N_{m,k}(t) = N_{m',k}(t)$ and $\hat{\mu}_{m,k}(t) = \hat{\mu}_{m',k}(t)$ for all $k=1,...,K$. Then, by construction of Selfish KL-UCB they will choose the same arm $\pi_{m}(t) = \pi_{m'}(t)$ and collide, and this cascade of collisions might go on forever.

We propose to alleviate the problem by adding randomization in order to break symmetry, by selecting arm
$$
    \pi_{m}(t) \in \arg\max_{k=1,...,K} \left\{ b_{m,k}(t)  +  {Z_{m,k}(t) \over t} \right\}
$$
where $(Z_{m,k}(t))_{m=1,...,M,k=1,...,K,t=1,...,T}$ are i.i.d. Gaussian with mean $0$ and variance $1$.

The variables $(Z_{m,k}(t))_{k=1,...,K}$ represent the internal randomization of player $m$ and it is noted that, of course, they are known only to player $m$. Informally, in order to break symmetry, each player maximizes the KL-UCB index perturbed by a small Gaussian perturbation. We call this algorithm Randomized Selfish KL-UCB, and we will show that it outperforms all known algorithms in Section~\ref{comparison_sota}. The pseudo code for Randomized Selfish KL-UCB is stated as Algorithm~\ref{alg:RandomizedSelfishKLUCB}.

\begin{figure}
    \centering
    \includegraphics[scale=0.5]{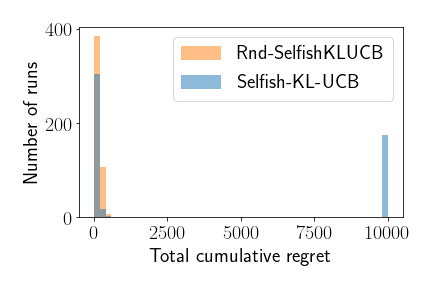}
    \includegraphics[scale=0.25]{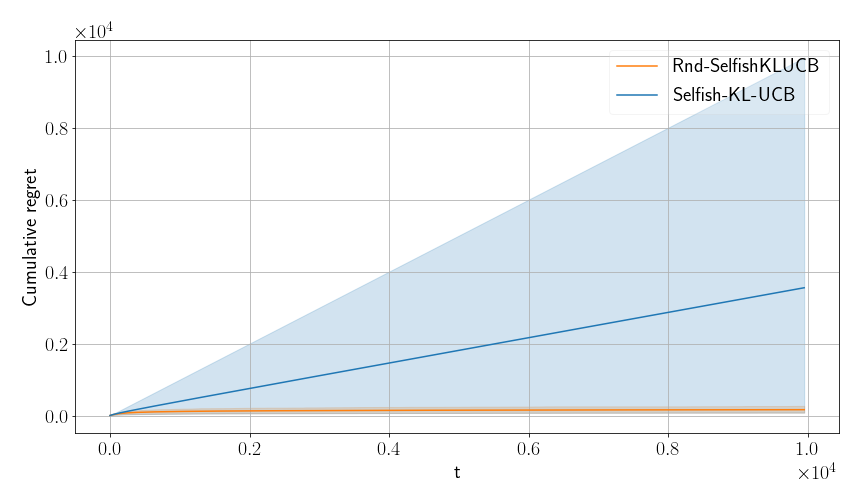}
    \caption{Environment: $M=2$, $K=2$, $\bm\mu = (0.9,0.1)$, $T=10,000$. Algorithms: Selfish KL-UCB, Randomized Selfish KL-UCB. Histogram of total cumulative regrets over 500 runs (left). Cumulative regret with respect to $t$, shaded areas represent the $90$-th percentile (right). }
    \label{fig:rnd_nornd}
\end{figure}

A rationale for Randomized Selfish KL-UCB is that, if two players have the same observations then while under Selfish KL-UCB they will choose the same arm with probability $1$ and trigger a potentially infinite cascade of collisions, under Randomized Selfish KL-UCB, there exists a positive probability that they will choose different arms, and collisions will eventually stop. The infinite cascade of collision phenomenon occurs especially often when the number of players and the number of arms are small. Figure~\ref{fig:rnd_nornd} illustrates this in an environment where $M=2$, $K=2$, that adding randomization allows to eliminate this problem: over $500$ runs, while Selfish KL-UCB incurs a linear regret for almost 200 runs, the histogram shows that there is only one mode for the total cumulative regret of Randomized Selfish KL-UCB, as it did not exceed $10^3$ in any of the runs.

\subsection{More Rationale for Selfish KL-UCB: Single Player Setting}

Another rationale for Randomized KL-UCB is understood by analyzing it in the single player case. Indeed, any good algorithm for the multi-player multi-armed bandit should at least be asymptotically optimal when applied in the single player case. Proposition~\ref{prop:singleplayer} states that Randomized Selfish KL-UCB, just like Selfish KL-UCB, is asymptotically optimal in this case.

By corollary, this also proves that Randomized Selfish KL-UCB performs well in a setting where a given player $m$ applies Randomized Selfish KL-UCB, while all other players select a constant arm, since this reduces to the single player case by replacing $\mu_k$ by $0$ if a player $m' \ne m$ plays arm $k$.

Also, when inspecting the proof in details, we understand why the magnitude of the randomization term is chosen as ${1 \over t}$, as it is small enough not to break asymptotic optimality at least in the single player case.
\begin{proposition}\label{prop:singleplayer}
    Consider the single player case $M=1$. Then under Randomized Selfish KL-UCB, for any $\bm\mu \in [0,1]^K$ and any $k$ such that $\mu_k < \mu_{(1)}$ we have
    $$
        \lim\sup_{T \to \infty} {\bE[N_{k}(T)] \over \log T} \le {1 \over d(\mu_k,\mu_{(1)}) } 
    $$
    i.e. the algorithm is asymptotically optimal.
\end{proposition}
Proof: see Section~\ref{sec:proofs}.

\subsection{Randomized Selfish KL-UCB: Analysis}

Despite our most sincere efforts, we were unable to prove a regret upper bound for Randomized Selfish KL-UCB in the multi-player setting. However, numerical experiments show that it outperforms all known algorithms sometimes by several orders of magnitude, as can be seen in the following section. We conjecture that Randomized Selfish KL-UCB has logarithmic regret, and we believe that this is an important, but certainly challenging open problem.

\section{Comparison to State-of-the-art Algorithms} \label{comparison_sota}

{\bf Algorithms} We now compare Randomized Selfish KL-UCB to state-of-the-art algorithms: EC-SIC \cite{shi_ec_sic}, SIC-MMAB2 \cite{boursier_sicmmab} and the algorithm 2 of \cite{lugosi} under various environments. For the settings with $M=2, K=3$, we also add the cooperative algorithm of \cite{bubeck}. Although it enjoys an asymptotic logarithmic regret, we do not plot the algorithm 1 of \cite{lugosi} because it converges too slowly: even for a very favorable case, such as with $\Delta = 0.5, M = 2, K = 3$ and $T = 2 \times 10^6$, the exploration phase lasts at least $24. 128K\log(3K M^2T^2)/ \Delta^2 > 1.2 \times 10^6$ time steps, leading to a linear regret in all our settings.

{\bf Parameter tuning} When algorithms require a hyperparameter depending on the environment, we input the best possible. That is, for SIC-MMAB2 and EC-SIC, that require a lower bound on $\mu_{(K)}$, we provide it with $\mu_{(K)}$. Similarly, for the algorithm 2 of \cite{lugosi} that we call Lugosi2 that requires a lower bound on $\mu_{(M)}$ we provide it with $\mu_{(M)}$. For EC-SIC, we use the parameter setting $p=5$ as suggested by \cite{shi_ec_sic}.

All experiments are averaged over at least $50$ runs, and the shaded areas represent $95\%$ confidence intervals.

\subsection{Linearly Spaced $\bf \mu$}
\begin{figure}[t]
\vskip -0.1in
\begin{center}
\includegraphics[width=0.32\columnwidth]{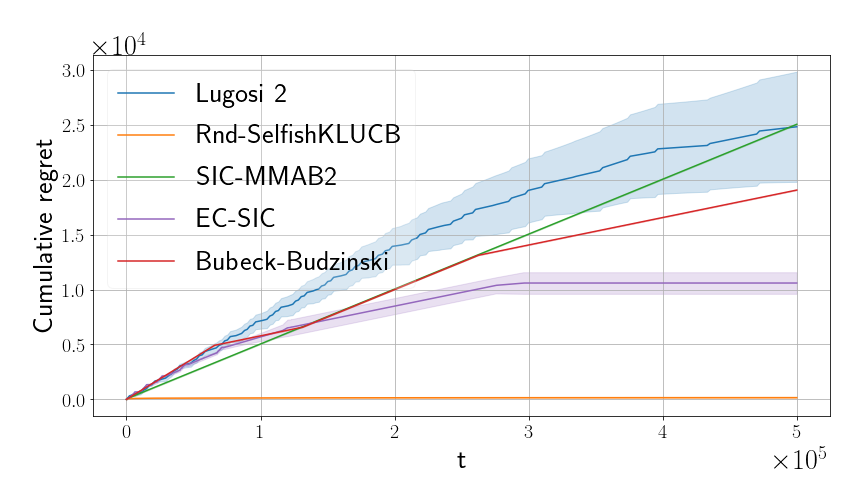}
\includegraphics[width=0.32\columnwidth]{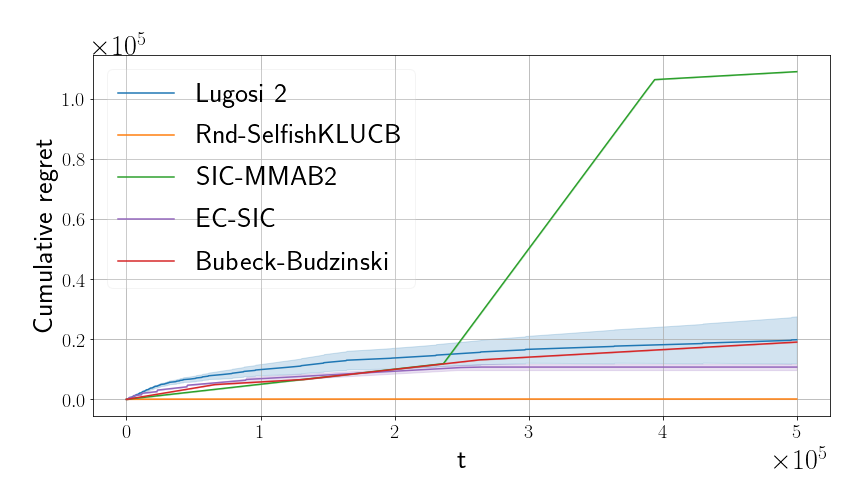}
\includegraphics[width=0.32\columnwidth]{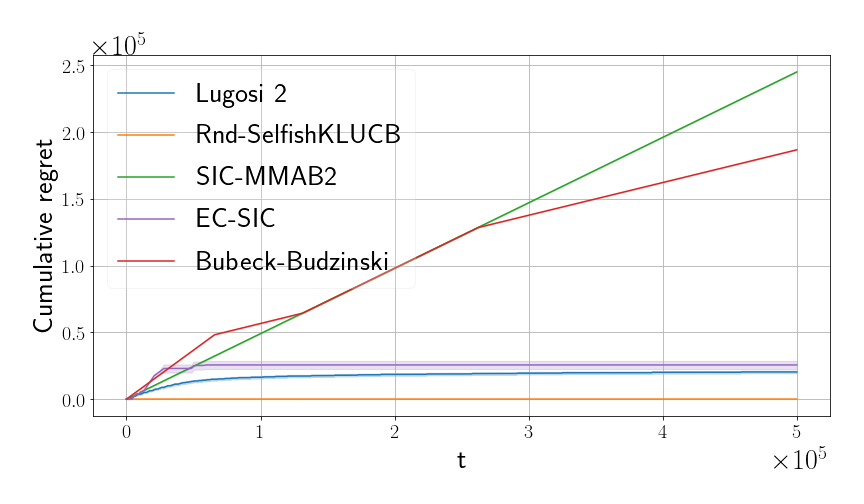}

\includegraphics[width=0.32\columnwidth]{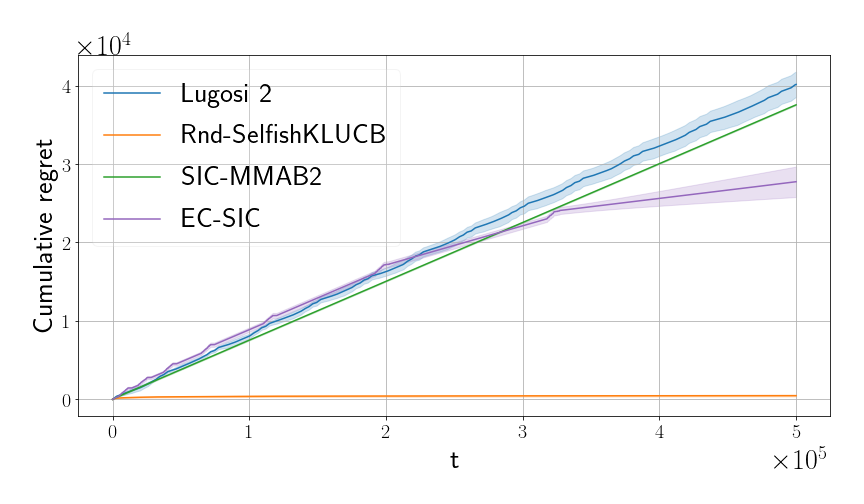}
\includegraphics[width=0.32\columnwidth]{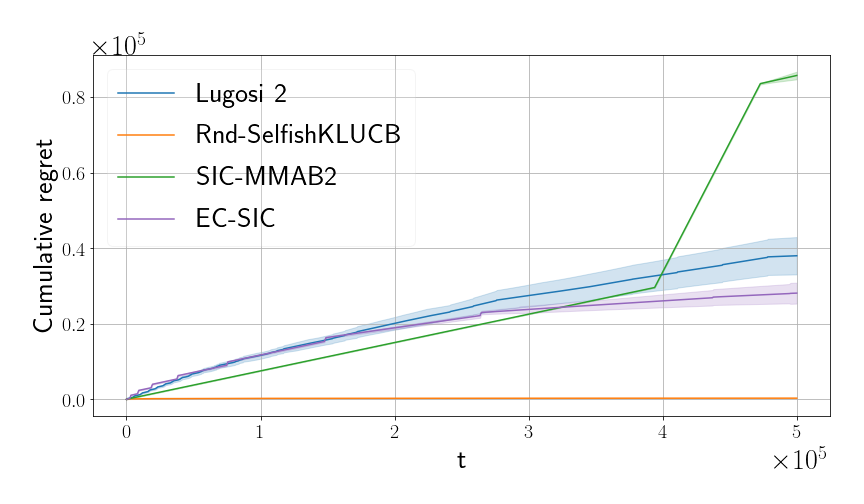}
\includegraphics[width=0.32\columnwidth]{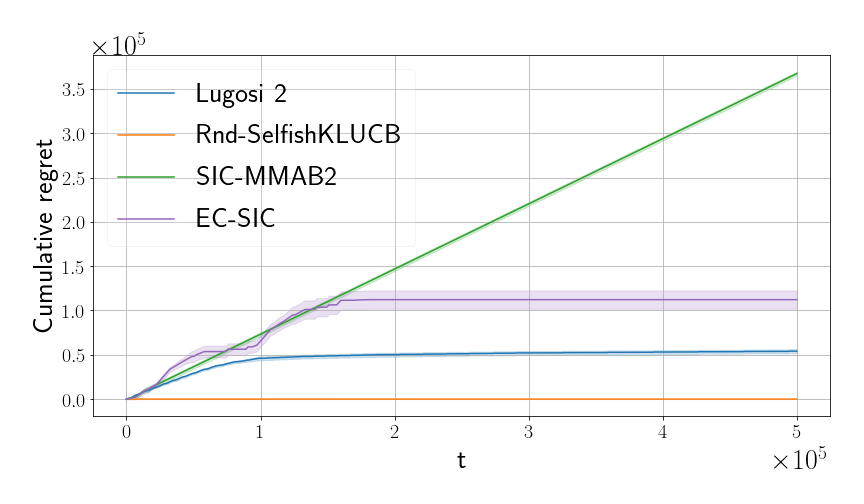}

\includegraphics[width=0.32\columnwidth]{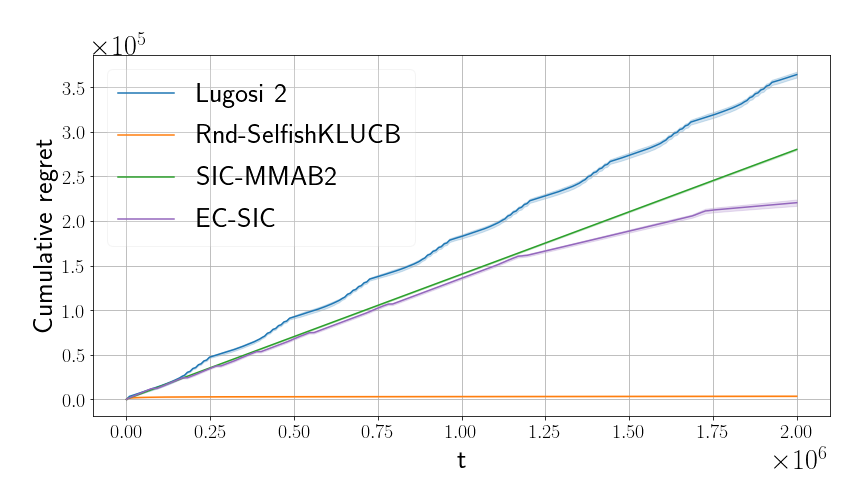}
\includegraphics[width=0.32\columnwidth]{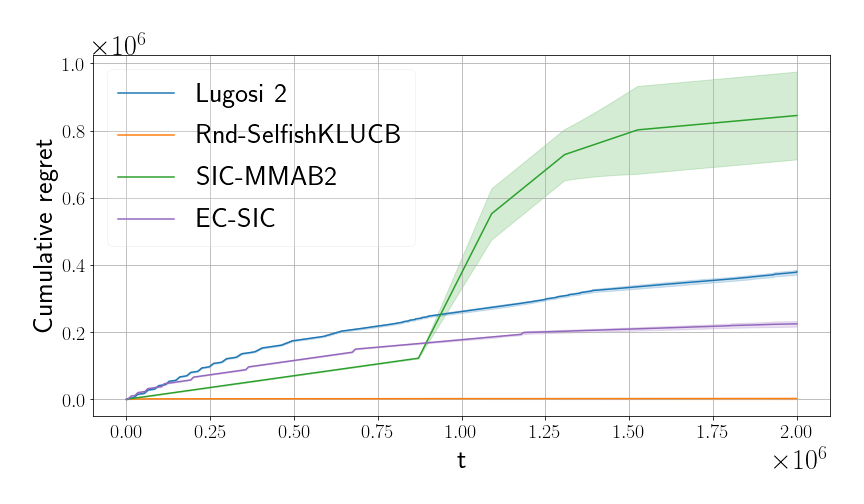}
\includegraphics[width=0.32\columnwidth]{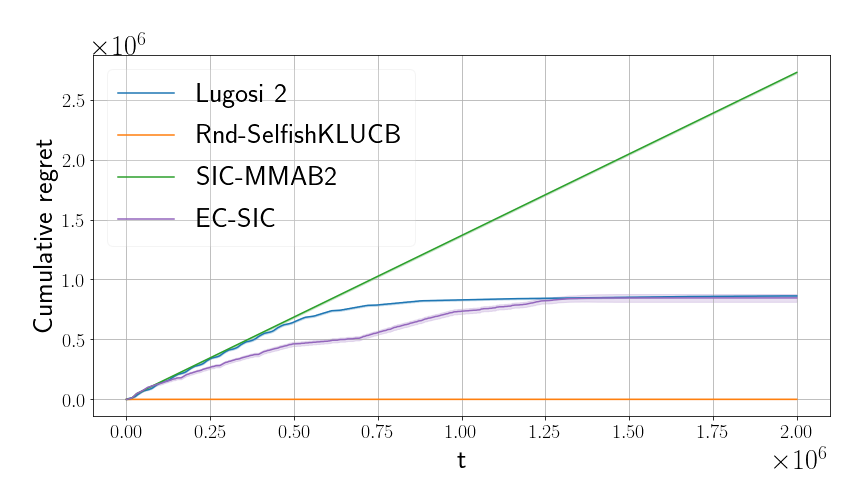}

\includegraphics[width=0.32\columnwidth]{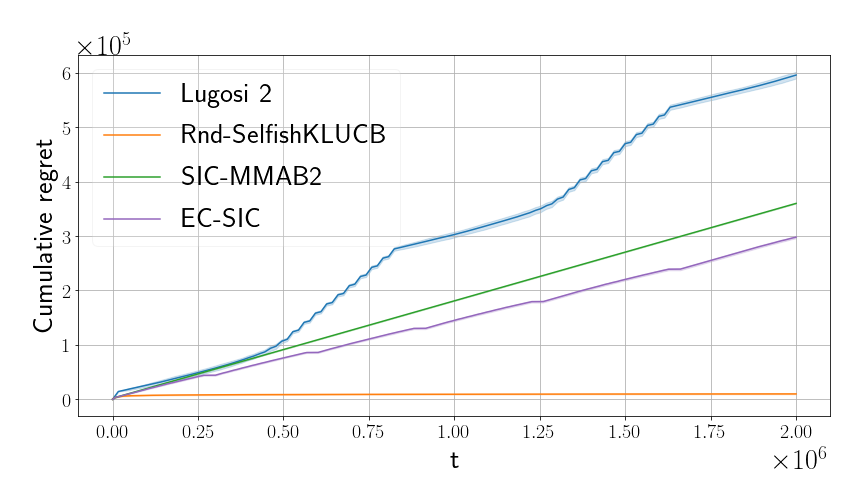}
\includegraphics[width=0.32\columnwidth]{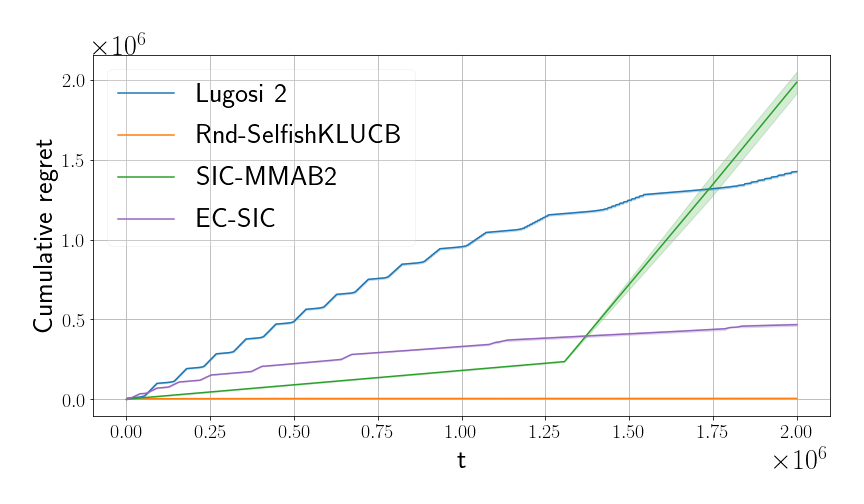}
\includegraphics[width=0.32\columnwidth]{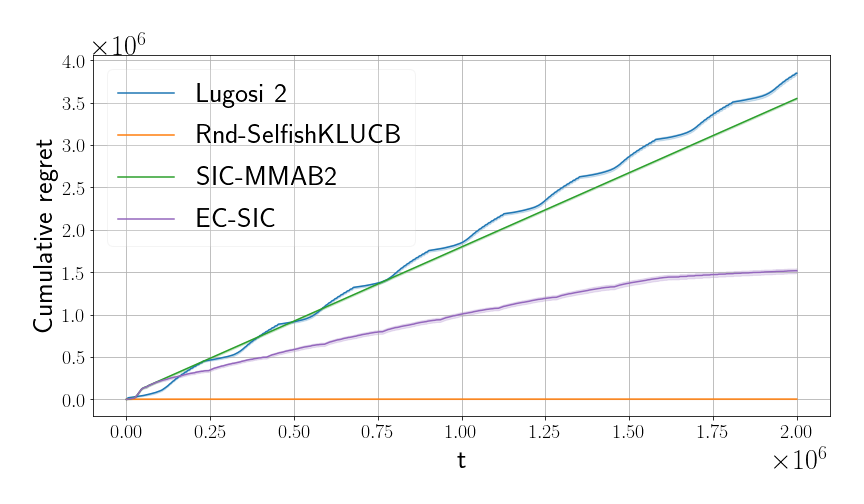}
\caption{Each row corresponds respectively to $(M,K) =(2,3)$, $(M,K) =(2,5)$, $(M,K) =(5,10)$, $(M,K)=(10,15)$. Each of the three columns corresponds respectively to an environment $\bm\mu$ generated by taking linearly spaced $\bm\mu = (0.2, \dots, 0.1)$, $\bm\mu = (0.9 , \dots,0.8)$, $ \bm\mu = (0.99, \dots,0.01). $}
\label{fig:linearly_m_2_K_3}
\end{center}
\vskip -0.2in
\end{figure}

We first evaluate algorithms on environments where the means of the arms are linearly spaced:
$$
    \mu_{(k)} = \mu_{(1)} {K-k \over K-1}  + \mu_{(K)} {k-1 \over K-1} \;,\; k=1,...,K
$$
We consider  $M =2, 5, 10$ players and a horizon of $T=2 \times 10^6$ time steps.
For each value of $(M,K)$, we evaluate the algorithms on three settings: \begin{itemize} \item{(i)} $(\mu_{(1)},\mu_{(K)})=(0.99,0.01)$ \item{(ii)} $(\mu_{(1)},\mu_{(K)})=(0.2,0.1)$ \item{(iii)} $(\mu_{(1)},\mu_{(K)})=(0.9,0.8)$ \end{itemize} We expect setting (i) to be the hardest and (iii) the easiest for SIC-MMAB2. Indeed, in setting (i) the length of its phases is large, for example when $(M,K) =(2,3)$, the first exploration phase has length $4800 K\log(T)/ \mu_{(K)} \ge 2 \times 10^7$, far greater than $T$ as illustrated in Figure~\ref{fig:linearly_m_2_K_3}. On the other hand in setting (iii) the first exploration phase lasts close to $3 \times 10^5$ time steps, and the following phases have a length of similar order. Although the phases length of EC-SIC are also inversely proportional to $\mu_{(K)}$, the communication of complete statistics of the players combined with the longer exploration phase ($p$ = 5) allow to classify the very good ($\mu_{(1)} = 0.99$) and very bad ($\mu_{(K)} = 0.01$) arms faster.

As shown by Figure~\ref{fig:linearly_m_2_K_3}, Randomized Selfish KL-UCB outperforms  other algorithms by far, and sometimes, by several orders of magnitude: for example, for $(M,K) =(5,10)$ in setting (ii), the regret of Randomized Selfish KL-UCB is $200$ times smaller than that of EC-SIC, the current best state-of-the-art algorithm.

\subsection{Variation of the Regret with Respect to Environment Parameters}
We now study how the variation of different quantities influence the algorithms performances.

We report the cumulative regret of each algorithm averaged over $50$ runs as a function of:
\begin{enumerate}
    \item $\mu_{(K)}$: for an environment with $M=5$ players and $K=9$ arms, we take $\bm\mu$ linearly spaced between $\mu_{(1)} =0.9$ and $\mu_{(K)}$, where $\mu_{(K)}$ varies between $0.1$ and $0.8$
    \item $\Delta$: for an environment with $M=5$ players and $K=9$ arms, we take $\bm\mu = (0.99, \dots, \mu_{(M)}, 0.8, \dots, 0.7)$, where $\mu_{(M)} \in \{0.9, 0.85, 0.81, 0.805,0.801\}$. Note that we chose $\mu_{(K)}$ high enough to favor the SIC algorithms, so that the regret of SIC-MMAB2 does not grow linearly like in the first column of Figure~\ref{fig:linearly_m_2_K_3}.
    \item $M$: for an environment with $K=10$ arms, we vary $M$ from $1$ to $9$ (EC-SIC does not work for $K=M$) and we take $\bm\mu$ linearly spaced between $\mu_{(1)} =0.9$ and $ \mu_{(K)} = 0.1$.
\end{enumerate}

\begin{figure}[ht]
\vskip -0.1in
\begin{center}
\includegraphics[scale=0.24]{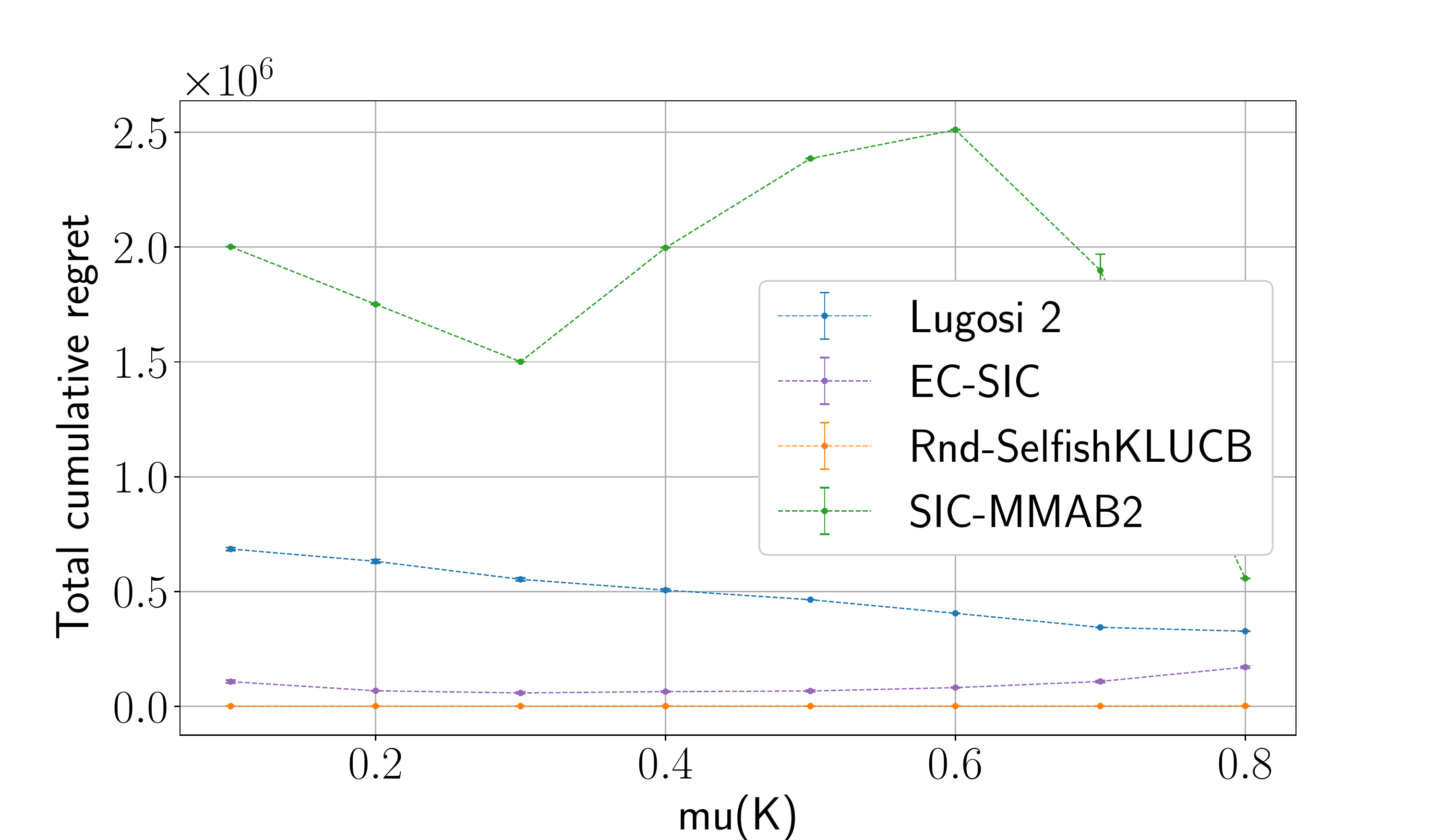}
\includegraphics[scale=0.24]{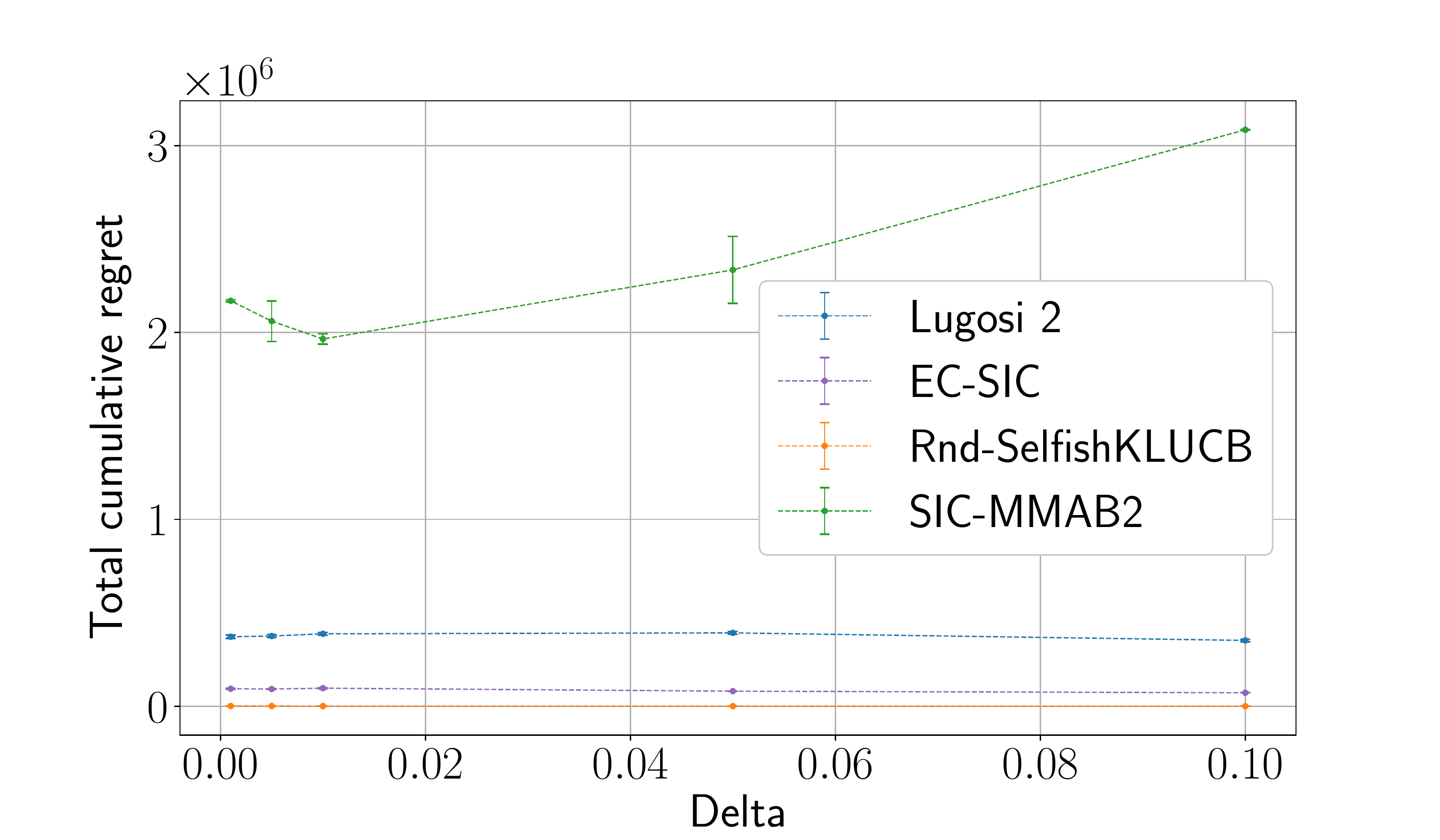}
\includegraphics[scale=0.24]{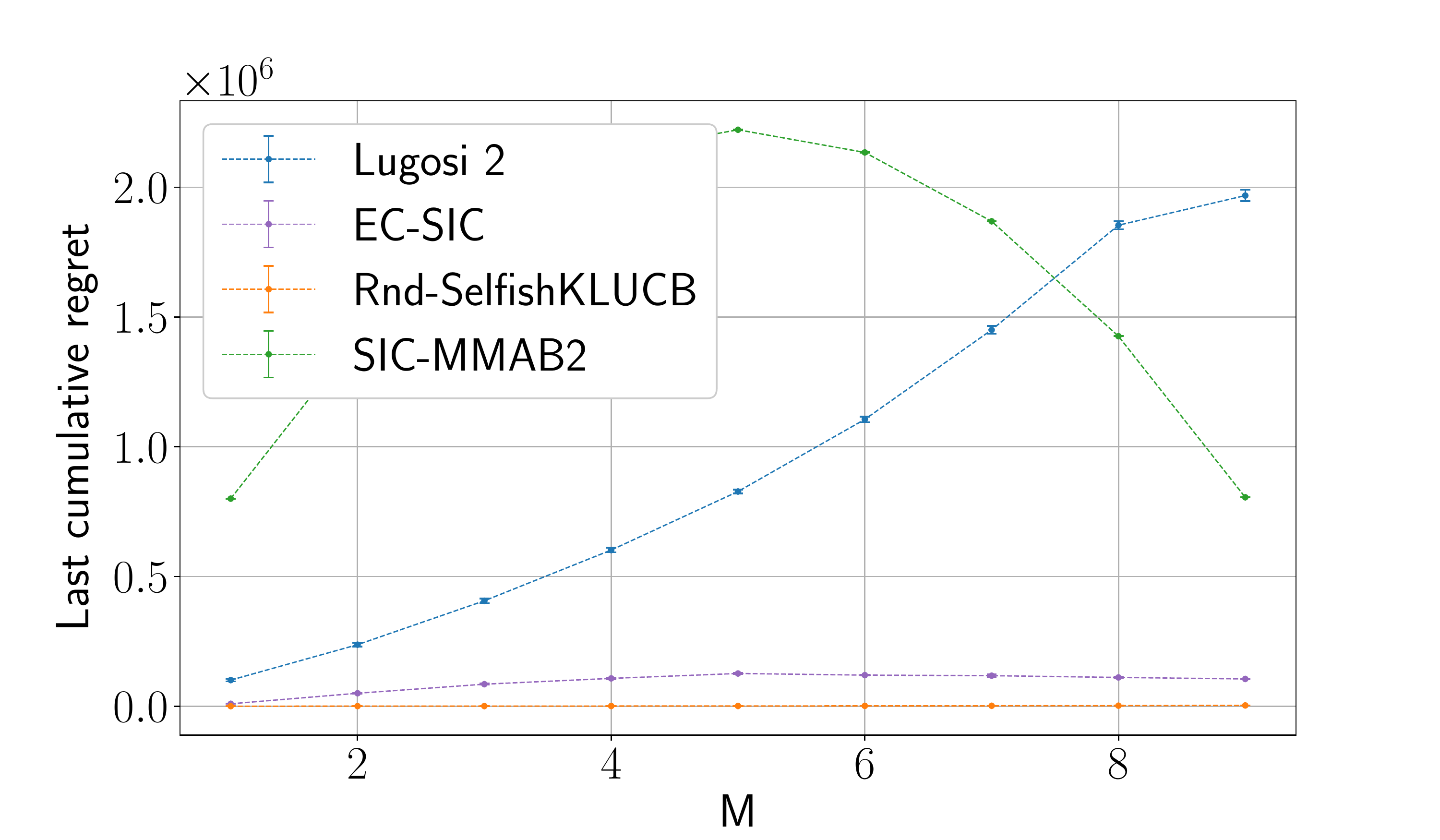}

\caption{Total cumulative regret as a function of $\mu_{(K)}$ (top), $\Delta$ (middle), $M$ (bottom) for all state-of-the-art algorithms.}\label{fig:wrt_delta_mu_min}
\end{center}
\vskip -0.2in
\end{figure}

For all these parameter values, Figure~\ref{fig:wrt_delta_mu_min} shows the superiority of Randomized Selfish KL-UCB over other algorithms. Note that even if EC-SIC seems close to Randomized Selfish KL-UCB in those three plots, the difference is actually quite significant.
Note also that although it seems that for small $\mu_{(K)}$, SIC-MMAB2 performs better than for very high $\mu_{(K)}$, it is actually not the case, because for small $\mu_{(K)}$, SIC-MMAB2 does not converge (similarly to the plots of the first column of Figure~\ref{fig:linearly_m_2_K_3}).

\subsection{A Corner Case: When All Means are Equal}

\begin{figure}[ht]
\vskip -0.1in
\begin{center}
\includegraphics[scale=0.25]{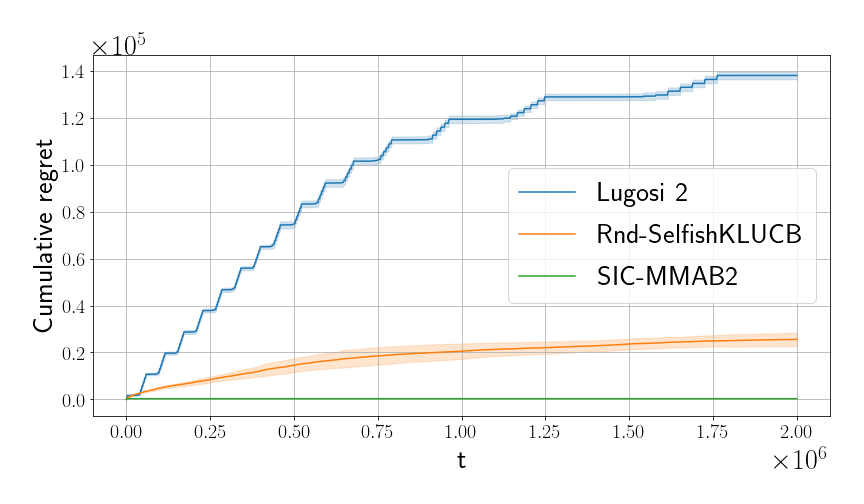}
\includegraphics[scale=0.25]{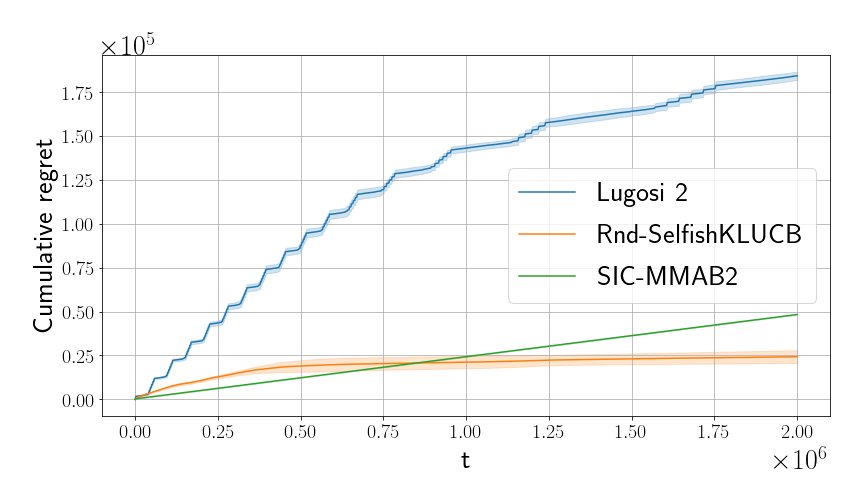}
\caption{$M=5, K=10$ When all means are equal $\bm\mu = (0.5, \dots, 0.5)$ (top), SIC-MMAB2 performs better than Randomized Selfish KL-UCB. When there is a small perturbation to this environment for example $\bm\mu \sim \cU([0.49,0.51]^K)$ (bottom), it is not the case anymore.}
\label{fig:same_mean}
\end{center}
\vskip -0.2in
\end{figure}

We found in the corner case where arms all have the same means reward, that Randomized Selfish KL-UCB does not perform better than SIC-MMAB, although its regret still seems logarithmic as can be seen in Figure~\ref{fig:same_mean}. This can be explained by the fact that when all arms have the same mean, the best strategy is just to always be in an orthogonalized setting, (and a simple Musical Chairs should actually be an optimal strategy) which is exactly how the SIC-MMAB2 algorithm behaves: players start with a Musical Chairs and then continue sequential hopping forever.

Nevertheless, this specific corner case is not likely to happen in practice, and the good performance of SIC-MMAB2 in this setting is not robust to even very small perturbations as shown in Figure~\ref{fig:same_mean}, where we added a noise sampled from a uniform distribution centered in $0.5$, of width $0.02$, so that $\bm\mu$ is uniformly distributed in $[0.49,0.51]^K$.

\section{Comparison to Algorithms Assuming Collision Sensing}\label{sec:comparison_wo_coll_info}
\begin{figure}[t]
\vskip -0.1in
\begin{center}
\includegraphics[width=0.32\columnwidth]{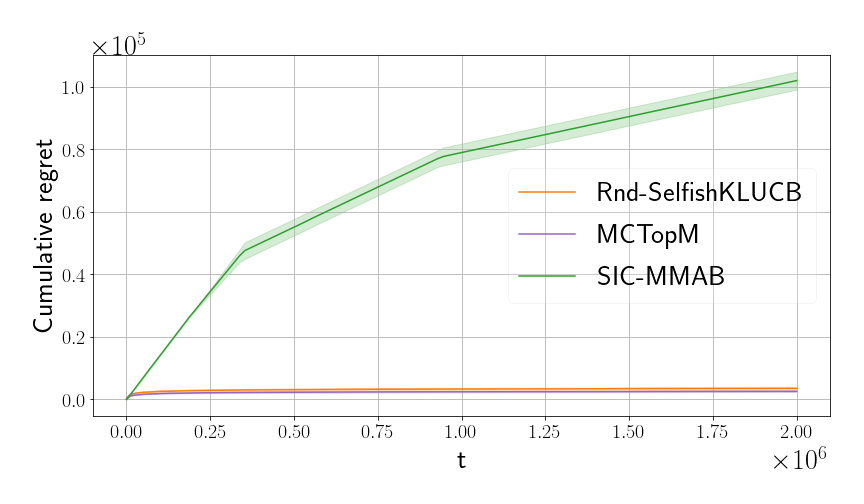}
\includegraphics[width=0.32\columnwidth]{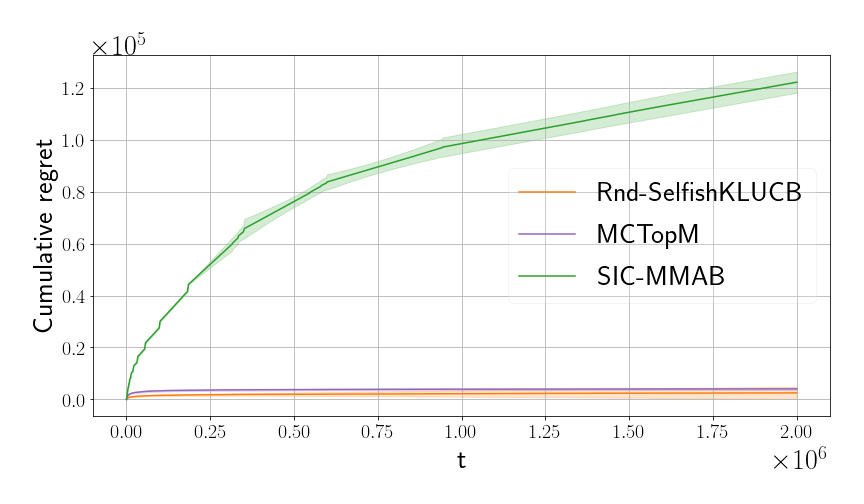}
\includegraphics[width=0.32\columnwidth]{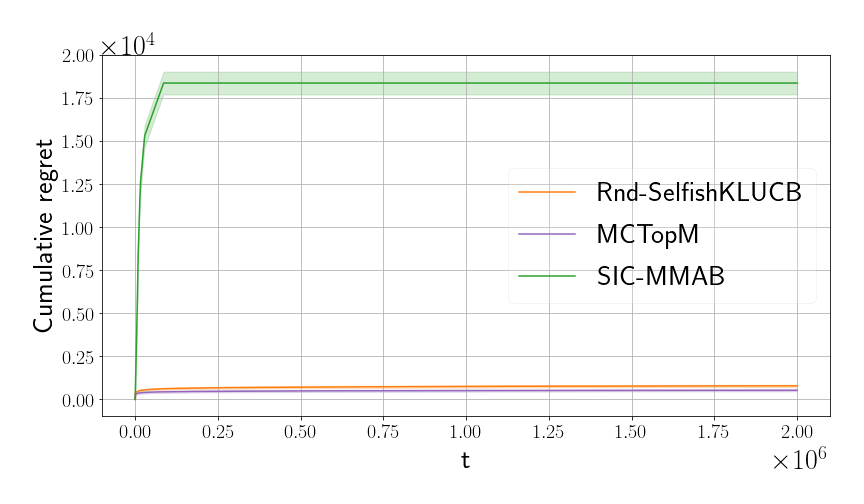}

\includegraphics[width=0.32\columnwidth]{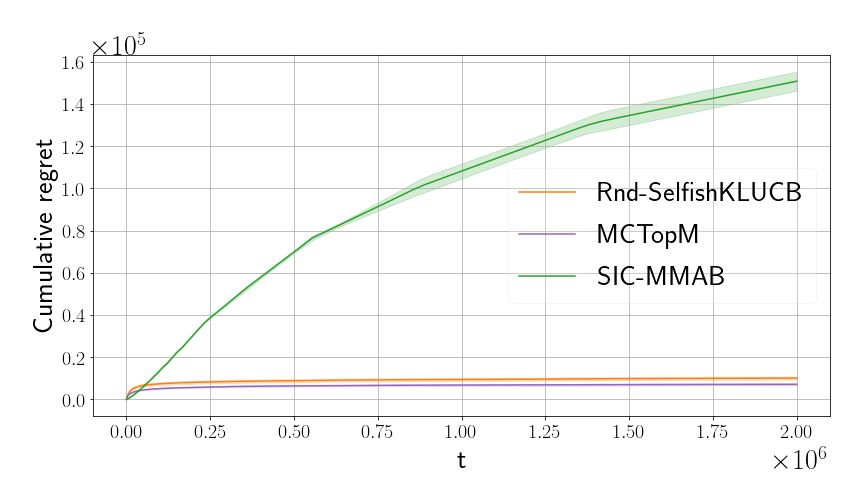}
\includegraphics[width=0.32\columnwidth]{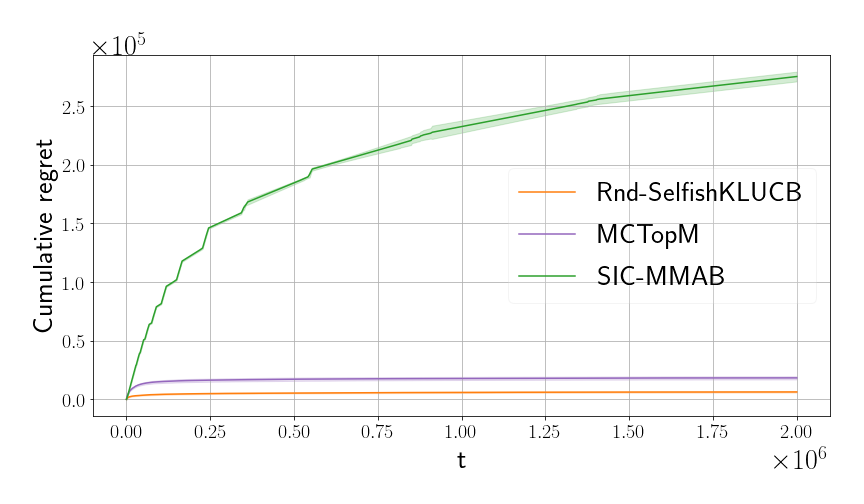}
\includegraphics[width=0.32\columnwidth]{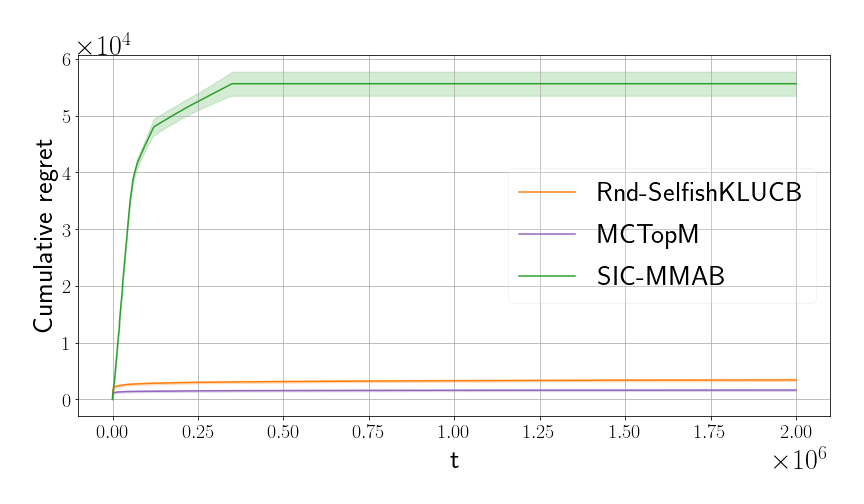}
\caption{Each row corresponds respectively to $(M,K) =(5,10)$ and $(M,K)=(10,15)$. Each of the three columns corresponds respectively to an environment $\bm\mu$ generated by taking linearly spaced $\bm\mu = (0.2, \dots, 0.1), \bm\mu = (0.9 , \dots,0.8),  \bm\mu = (0.99, \dots,0.01). $}
\label{fig:comparison_with_coll_info}
\end{center}
\vskip -0.2in
\end{figure}

In this section, we compare Randomized Selfish KL-UCB to algorithms which are state-of-the-art under the setting with collision and/or sensing information: SIC-MMAB \citep{boursier_sicmmab} and MCTopM \citep{besson_multiplayer_revisited}. It is interesting to see that Randomized Selfish KL-UCB often performs far better than SIC-MMAB as shown by Figure~\ref{fig:comparison_with_coll_info}, and its performance approaches that of MCTopM, sometimes outperforming it in certain environments. 

\section{Dynamical Setting}
As was noted by \cite{boursier_sicmmab}, the SIC algorithms (SIC-MMAB, SIC-MMAB2 and EC-SIC) rely heavily on the static assumption, so as to allow communication between players and "hack the system", by using a perfect synchronization between all players at all times. In practical applications however, this assumption is very unrealistic as players do not arrive at the same time, and do not leave at the same time (for instance in communication networks).

In this section, we study experimentally the dynamical setting without collision nor sensing information, a setting where little work has been done so far. 

\subsection{Quasi-Asynchronicity}
For the dynamical setting without collision information, \cite{boursier_sicmmab} proposed DYN-MMAB an algorithm with logarithmic regret. They consider the quasi-asynchronous setting, where players can enter the system whenever they want, but cannot leave until the end of the time horizon.

Formally, player $m$ enters at time $\tau_m \in \{0,...,T\}$ and
stays until the final horizon $T$. The value of $\tau_m$ is unknown to all players (including $m$), who are only aware of their individual horizon $T-\tau_m$ and their own internal clock $t - \tau_m$.

We model the arrival of players by a Poisson process, starting with one player at the beginning of the game. We let the maximum number of players be $M=K$, therefore for sufficiently long horizon the system ends up saturated at the end of the game.

\begin{figure}[h]
\begin{center}
\includegraphics[scale=0.45]{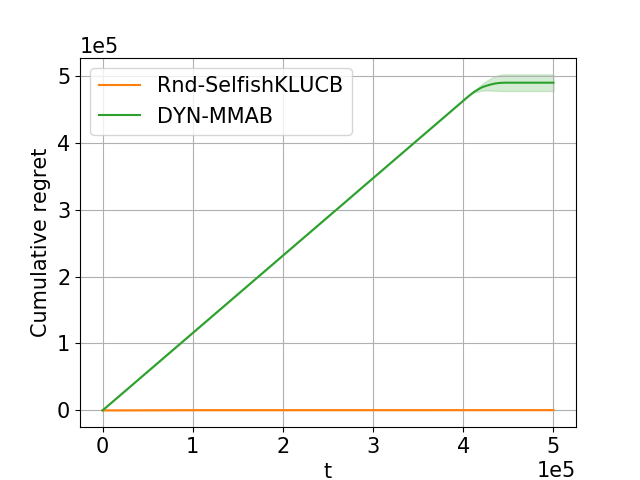}
\includegraphics[scale=0.45]{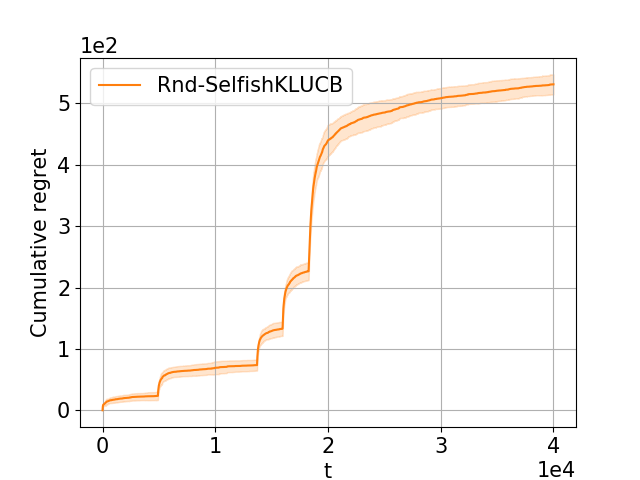}
\caption{Quasi-asynchronous environment with $\lambda = 10^{-4}$. Cumulative regret of DYN-MMAB and Randomized Selfish KL-UCB $K=4$ (left). Cumulative regret of Randomized Selfish KL-UCB, with $M=5$, players entering at $ \{0,  4912, 13703, 15970, 18278\}$ (sample from a Poisson process with rate $\lambda = 10^{-4}$) (right).}
\label{fig:dynamic}
\end{center}
\vskip -0.2in
\end{figure}

Figure~\ref{fig:dynamic} shows that DYN-MMAB converges slowly in comparison to Randomized Selfish KL-UCB. For Randomized Selfish KL-UCB, this setting might actually be even easier as players enter sequentially. Intuitively, if $M$ players have been playing for a long time in the game, they likely have settled on a preferred arm. If a new player enters, she effectively faces a system akin to a single player bandit with $K-M$ arms. This especially makes sense in light of proposition~\ref{prop:singleplayer} which treats the single-player case.

\subsection{When Players are Allowed to Leave}
Although the quasi-asynchronous setting is a step forward towards a realistic dynamical setting, in many real-world systems users leave whenever they want, in an asynchronous manner.

If we add the assumption that players can only leave at specific intervals, \cite{boursier_sicmmab} propose to adapt DYN-MMAB by resetting the algorithm at each of these intervals. 

We propose to study experimentally an even more realistic dynamic setting in which players can enter and leave the system at any moment. Denote by $M(t)$ the number of players present in the system at time $t$. We model the arrivals and departures as an M/M/K queue:
\begin{itemize}
    \item arrivals follow a Poisson process with rate $\lambda$,
    \item players stay for an exponentially distributed duration with mean $1/\nu$,
    \item when the system is saturated, that is $M(t)=K$, entering players are blocked.
\end{itemize}
As players constantly enter and leave the system, instead of reporting the cumulative regret with respect to $T$, we measure the performance of Randomized Selfish KL-UCB by computing an expected reward per unit time $\cR$. More specifically, the performance of an algorithm is 
\begin{align*} \cR &= {1 \over T} \sum_{t=1}^T \sum_{m=1}^{M(t)} \bE [r_{\pi_m}(t)] \\
&= {1 \over T}  \sum_{t=1}^T \sum_{m=1}^{M(t)} \bE \Big[\mu_{\pi_m}(t) (1 -\eta_{\pi_m(t)}(t))\Big],
\end{align*} 
while the performance of the optimal oracle algorithm is 
$$\cR^\star = \frac{1}{T} \sum_{t=1}^T \sum_{m=1}^{M(t)} \mu_{(m)}$$
and we report the performance ratio ${\cR \over \cR^\star}$. 

Under the realistic scenario where players arrive at a rate of $1$ person per second (where $1$ second corresponds to $10^3$ time steps), and stay for $10$ seconds ($10^4$ time steps), with an average number of $8$ players, the algorithm achieves a ratio of $95\%$ compared to the optimal oracle algorithm. 

We report in Table~\ref{table:dyn_mc}, the ratio between the performance of Randomized Selfish KL-UCB and that of the optimal oracle algorithm, with respect to multiple parameters of $\lambda$ and $\nu$.

With this model, Randomized Selfish KL-UCB performs almost like the optimal oracle algorithm. 
In comparison, we also report the ratio between the performance of Musical Chairs and that of the optimal oracle algorithm.

\begin{table}[h]
  \caption{Dynamic setting. Performance ratio against optimal oracle algorithm, for Randomized Selfish KL-UCB (top) and Musical Chairs (bottom), with $T = 10^6$, $K=10$. }
  \label{table:dyn_mc}
\centering
\begin{small}
    \begin{tabular}{|l|c|c|r|}
      \hline
      Randomized Selfish KL-UCB     & $\lambda = 1/1000$ & $\lambda = 1/10,000$  \\
      \hline
      $\nu = 1/500$ & 91 $\pm$ 2 \% & 91 $\pm$ 1 \%  \\
      \hline
      $\nu = 1/1000$ & 92  $\pm$ 2\% & 94 $\pm$ 1 \% \\
      \hline
      $\nu = 1/10,000$ & 93 $\pm$ 1 \% & 97 $\pm$ 1 \% \\
      \hline
    \end{tabular}
    
    \medskip
    
    \begin{tabular}{|l|c|c|r|}
      \hline
      Musical Chairs     & $\lambda = 1/1000$ & $\lambda = 1/10,000$  \\
      \hline
      $\nu = 1/500$ & 69 $\pm$ 1 \% & 69 $\pm$ 3\%  \\
      \hline
      $\nu = 1/1000$ & 72  $\pm$ 1\% & 70 $\pm$ 3 \% \\
      \hline
      $\nu = 1/10,000$ & 90 $\pm$1 \% & 72 $\pm$ 3 \% \\
      \hline
    \end{tabular}
  
\end{small}

\end{table}

\section{Proof of Proposition~\ref{prop:singleplayer}}\label{sec:proofs}
\subsection{Technical Results}
\begin{lemma}[Chernoff bound for Gaussian variables] \label{lem:chenoff_gaussian}
    Consider $Z \sim N(0,\sigma^2)$. Then for all $\delta > 0$ we have $\bP( Z \ge \delta) \le e^{- {\delta^2 \over 2 \sigma^2}}$.
\end{lemma}
{\bf Proof:} A Chernoff bound yields, for any $\lambda > 0$
\begin{align*}
    \bP( Z \ge \delta) =  \bP( e^{\lambda Z} \ge e^{\lambda \delta}) \le e^{-\delta \lambda }\bE e^{\lambda Z} = e^{ {\lambda^2 \sigma^2 \over 2} -\delta \lambda }
\end{align*}
and setting $\lambda = \sigma^2/\delta$ yields the result.

\subsection{Proof of Proposition~\ref{prop:singleplayer}}

Consider the single-player case $M=1$. Define $k^\star \in \arg\max_{k=1,...,K} \mu_k$ an optimal arm, consider $k$ a suboptimal arm so that $\mu_k < \mu_{k^\star}$. Consider $0 < \delta < (\mu_{k^\star} - \mu_k)/2$ fixed. The analysis is based on that of KL-UCB  \cite{klucb}.

Define the following events:
\begin{align*}
    {\cal A}_t &= \{ b_{k^\star}(t) \le \mu_{k^\star} \} \\
    {\cal B}_t &= \{ Z_{k}(t) - Z_{k^\star}(t) \ge t \delta \} \\
    {\cal C}_t &= \{ k(t) = k: N_k(t) \le f(T)/d(\mu_k+\delta,\mu_{k^\star} - \delta)  \} \\
    {\cal D}_t &= \{ k(t) = k: \hat{\mu}_k(t) \ge  \mu_k + \delta  \}.
\end{align*}

Let us prove that if none of those events occur, then $k(t) \ne k$ i.e. $k$ cannot be selected. If ${\cal A}_t$ does not occur then $b_{k^\star}(t) \ge \mu_{k^\star}$. If ${\cal C}_t$ and ${\cal D}_t$ both do not occur we have:
\begin{align*}
    N_k(t) d(\hat{\mu}_k(t),\mu_{k^\star} - \delta) &\ge N_k(t) d(\hat{\mu}_k(t)-\delta,\mu_{k^\star} - \delta) \ \\
    &\ge f(T) \ge f(t)
\end{align*}
therefore $b_k(t) \le \mu_{k^\star} - \delta$. If ${\cal B}_t$ does not occur as well, we finally get
$$
    b_k(t) + {Z_k(t) \over t} \le b_{k^\star}(t) + {Z_{k^\star}(t) \over t} 
$$
so that indeed, we cannot have $k(t) = k$.

So the number of times $k$ is selected is upper bounded as:
$$
 N_{k}(T) \le \sum_{t=1}^T \indic\{ {\cal A}_t \} + \indic\{ {\cal B}_t \} + \indic\{ {\cal C}_t \} + \indic\{ {\cal D}_t \}
$$
From \cite{klucb} we have that:
$$
    \sum_{t=1}^{+\infty} \bP({\cal A}_t) \le C_1 \log \log T
$$
with $C_1 \ge 0$ a universal constant. Using the fact that $Z_k(t) - Z_{k}(t)$ has $N(0,2)$ distribution,  using lemma~\ref{lem:chenoff_gaussian}
$$
    \sum_{t=1}^{+\infty} \bP({\cal B}_t) \le \sum_{t=1}^{+\infty} e^{- t^2 \delta^2/2 } < +\infty
$$
When ${\cal C}_t$ occurs we have that $N_k$ is incremented so that from a counting argument
$$
    \sum_{t=1}^{+\infty} \indic\{ {\cal C}_t \} \le {f(T) \over d(\mu_k+\delta,\mu_{k^\star} - \delta)}
$$  
Finally, using Hoeffding's inequality:
$$
   \sum_{t=1}^{+\infty} \bP({\cal D}_t) \le  \sum_{n=0}^{+\infty} e^{- 2 n \delta^2} = {1 \over 1 -  \exp^{- 2 \delta^2}} < \infty
$$
Putting it together we have proven that
$$
    \lim\sup_{T \to \infty} {\bE[N_{k}(T)] \over \log T} \le {1 \over d(\mu_k+\delta,\mu_{k^\star} - \delta)}
$$
Since the above holds for $\delta$ arbitrarily small we have proven the announced result:
$$
    \lim\sup_{T \to \infty} {\bE[N_{k}(T)] \over \log T} \le {1 \over d(\mu_k,\mu_{k^\star})}.
$$

\section{Conclusion}

In this work, through extensive experiments, we emphasize the potential of Randomized Selfish KL-UCB as an optimal algorithm for the decentralized MP-MAB without collision and sensing information for the static setting. We argue that for real-world applications, Randomized Selfish KL-UCB is a very good candidate as it performs well, does not require any prior knowledge on the environment, and is simple to implement in comparison to its peers which rely on complex multiple phases and sometimes unrealistic communication through collisions between users. Moreover, for the more realistic dynamic setting, our experiments also show promising results.
We hope this work will encourage the community toward the analysis of this algorithm, a challenging but promising open problem. 
 


\begin{algorithm}[tb]
   \caption{Selfish KL-UCB (for player $m=1,...,M$)}
   \label{alg:SelfishKLUCB}
\begin{algorithmic}

   \FOR{$t=1, \dots, T$}
        \FOR{$k=1,...,K$}
        \STATE Compute 
        $
    b_{m,k}(t) = \max \{q \in [0,1]: N_{m,k}(t) d(  \hat{\mu}_{m,k}(t), q ) \le f(t) \}
$      \ENDFOR
   \STATE  Draw arm $\pi_m(t) = \argmax{k=1,...,K} \ \{ b_{m,k}(t) \}$ 
    \STATE  Observe reward $r_m(t)$ and update statistics\;
   \ENDFOR
\end{algorithmic}
\end{algorithm}

\begin{algorithm}[tb]
   \caption{Randomized Selfish KL-UCB (for player $m=1,...,M$)}
   \label{alg:RandomizedSelfishKLUCB}
\begin{algorithmic}

   \FOR{$t=1, \dots, T$}
        \FOR{$k=1,...,K$}
        \STATE Compute 
        $
    b_{m,k}(t) = \max \{q \in [0,1]: N_{m,k}(t) d(  \hat{\mu}_{m,k}(t) , q ) \le f(t) \}$
        \STATE Draw $Z_{m,k}(t) \sim N(0,1)$
      \ENDFOR
   \STATE  Draw arm $\pi_m(t) = \argmax{k=1,...,K} \ \{ b_{m,k}(t) + {Z_{m,k}(t) \over t}\}$ 
    \STATE  Observe reward $r_m(t)$ and update statistics\;
   \ENDFOR
\end{algorithmic}
\end{algorithm}

\bibliography{references}
\bibliographystyle{plainnat}
\setcitestyle{authoryear,round,citesep={;},aysep={,},yysep={;}}

\end{document}